\RequirePackage{snapshot}

\documentclass{article}

\usepackage{microtype}
\usepackage{graphicx}
\usepackage{subcaption}
\usepackage{booktabs} % for professional tables
\usepackage{verbatim}
\usepackage{soul}
\usepackage{amsmath}
\usepackage{amsthm}
\usepackage{amsfonts}
\usepackage{amssymb}
\usepackage{color}
\usepackage{lipsum}
\usepackage{enumitem}
\usepackage{microtype}
\usepackage{marginnote}
\usepackage{caption}

\newcommand{\GNAT}{GLO}

\usepackage{hyperref}

\usepackage[accepted]{icml2018}

\icmltitlerunning{Optimizing the Latent Space of Generative Networks}

\begin{document}

\twocolumn[

\icmltitle{Optimizing the Latent Space of Generative Networks}



\icmlsetsymbol{equal}{*}

\begin{icmlauthorlist}
\icmlauthor{Piotr Bojanowski}{fb}
\icmlauthor{Armand Joulin}{fb}
\icmlauthor{David Lopez Paz}{fb}
\icmlauthor{Arthur Szlam}{fb}
\end{icmlauthorlist}

% \author{Piotr Bojanowski, Armand Joulin, David Lopez-Paz, Arthur Szlam\\
% Facebook AI Research\\
% \texttt{\{bojanowski, ajoulin, dlp, aszlam\}@fb.com}}

\icmlaffiliation{fb}{Facebook AI Research}

\icmlcorrespondingauthor{Piotr Bojanowski}{bojanowski@fb.com}

\icmlkeywords{Generative Models, Unsupervised Learning}

\vskip 0.3in
]



\printAffiliationsAndNotice{}  % leave blank if no need to mention equal contribution

\begin{abstract}
Generative Adversarial Networks (GANs) have achieved remarkable results in the task of generating realistic natural images.
In most successful applications, GAN models share two common aspects:
 solving a challenging saddle point optimization problem, interpreted as an adversarial game between a generator and a discriminator functions;
and parameterizing  the generator and the discriminator as  deep convolutional neural networks.
The goal of this paper is to disentangle the contribution of these two factors to the success of GANs.
In particular, we introduce \emph{Generative Latent Optimization}~(\GNAT{}), a framework to train deep convolutional generators using simple reconstruction losses.
Throughout a variety of experiments, we show that \GNAT{} enjoys many of the desirable properties of GANs: synthesizing visually-appealing samples, interpolating meaningfully between samples, and performing linear arithmetic with noise vectors; all of this without the adversarial optimization scheme.
\end{abstract}

\section{Introduction}\label{sec:introduction}
% description of the principle of GAN training
Generative Adversarial Networks~ (GANs)~\citep{gan} are a powerful framework to learn models capable of generating natural images.
GANs learn these generative models by setting up an adversarial game between two learning machines.
On the one hand, a {generator} plays to transform {noise vectors} into {fake samples}, which resemble {real samples} drawn from a distribution of natural images.
On the other hand, a {discriminator} plays to distinguish between real and fake samples.
During training, the generator and the discriminator functions are optimized in turns.
First, the discriminator learns to assign high scores to real samples, and low scores to fake samples.
Then, the generator learns to increase the scores of fake samples, so as to ``fool'' the discriminator.
After proper training, the generator is able to produce realistic natural images from noise vectors.

% features of GANs that show that they generalize: interpolation, arithmetic, and sampling
Recently, GANs have been used to produce high-quality images resembling handwritten digits, human faces, and house interiors~\citep{dcgan}.
Furthermore, GANs exhibit three strong signs of generalization.
First, the generator translates~\emph{linear interpolations in the noise space} into~\emph{semantic interpolations in the image space}.
In other words, a linear interpolation in the noise space will generate a smooth interpolation of visually-appealing images.
Second, the generator allows \emph{linear arithmetic in the noise space}.
Similarly to word embeddings \citep{mikolov2013efficient}, linear arithmetic indicates that the generator organizes the noise space to disentangle the nonlinear factors of variation of natural images into linear statistics.
Third, the generator is able to to synthesize new images that resemble those of the data distribution.
This allows for applications such as image in-painting \citep{IizukaSIGGRAPH2017} and super-resolution~\citep{ledig2016photo}.

% hard training of GANS
Despite their success, training and evaluating GANs is notoriously difficult.
The adversarial optimization problem implemented by GANs is sensitive to random initialization, architectural choices, and hyper-parameter settings.
In many cases, a fair amount of human care is necessary to find the correct configuration to train a GAN in a particular dataset.
It is common to observe generators with similar architectures and hyper-parameters to exhibit dramatically different behaviors.
Even when properly trained, the resulting generator may synthesize samples that resemble only a few localized regions (or modes) of the data distribution~\citep{gan_tutorial}.
While several advances have been made to stabilize the training of GANs~\citep{salimans2016improved}, this task remains more art than science.

% it is not only hard to train such models, it is also quite hard to evaluate
The difficulty of training GANs is aggravated by the challenges in their evaluation: since evaluating the likelihood of a GAN with respect to the data is an intractable problem, the current gold standard to evaluate the quality of GANs is to eyeball the samples produced by the generator.
This qualitative evaluation gives little insight on the coverage of the generator, making the mode dropping issue hard to measure.
The evaluation of discriminators is also difficult, since their visual features do not always transfer well to supervised tasks~\citep{donahue2016adversarial,dumoulin2016adversarially}.
Finally, the application of GANs to non-image data has been relatively limited.

%%%%%%%%%%%%%%%%%%%%%%%%%
%%% PROBLEM STATEMENT %%%
%%%%%%%%%%%%%%%%%%%%%%%%%

\subsection{Research question}
To model natural images with GANs, the generator and discriminator are commonly parametrized as deep Convolutional Networks~(convnets)~\citep{lecun1998gradient}.
Therefore, it is reasonable to hypothesize that the reasons for the success of GANs in modeling natural images come from two complementary sources:
\begin{enumerate}[label=(A\arabic*),start=1]
  \item Leveraging the powerful inductive bias of deep convnets.
  \item The adversarial training protocol.
\end{enumerate}
This work attempts to disentangle the factors of success (A1) and (A2) in GAN models.
Specifically, we propose and study one algorithm that relies on (A1) and avoids (A2), but still obtains competitive results when compared to a GAN.

\paragraph{Contributions.}
We investigate the importance of the inductive bias of convnets by removing the adversarial training protocol of GANs (Section~\ref{sec:gnat}).
Our approach, called \emph{Generative Latent Optimization}~(\GNAT{}), maps one \emph{learnable} noise vector to each of the images in our dataset by minimizing a simple reconstruction loss.
Since we are predicting~\emph{images from learnable noise},~\GNAT{} borrows inspiration from recent methods to predict~\emph{learnable noise from images}~\citep{nat}.
Alternatively, one can understand~\GNAT{} as an auto-encoder where the latent representation is not produced by a parametric encoder, but learned freely in a non-parametric manner.
In contrast to GANs, we track the correspondence between each learned noise vector and the image that it represents.
Hence, the goal of \GNAT{} is to find a meaningful organization of the noise vectors, such that they can be mapped to their target images.
To turn \GNAT{} into a generative model, we observe that it suffices to learn a simple probability distribution on the learned noise vectors.

We study the efficacy of \GNAT{} to compress and decompress a dataset of images, generate new samples, perform linear interpolations and extrapolations in the noise space, and perform linear arithmetic.
Our experiments provide quantitative and qualitative comparisons to Principal Component Analysis (PCA), Variational Autoencoders (VAE) and GANs.    Our results show that on many image datasets, in particular CelebA, MNIST and SVHN, the celebrated properties of GAN generations can be reproduced without the GAN training protocol.
On the other hand, our qualitative results on the LSUN bedrooms are worse than the results of GANs; we hypothesize (and show evidence) that this is a capacity issue.
It has been observed that GANs are prone to mode collapse, completely forgetting large parts of the training dataset.
In the literature this is often described as a problem with the GAN training procedure.
Our experiments suggest that this is more of a feature than a bug, as it allows relatively small models to generate realistic images by intelligently choosing which part of the data to ignore.
We quantitatively measure the significance of this issue with a reconstruction criterion.

\begin{figure}[t]
  \centering
  \includegraphics[width=0.99\linewidth]{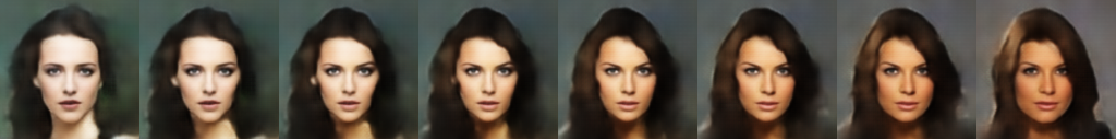} \\
  \includegraphics[width=0.99\linewidth]{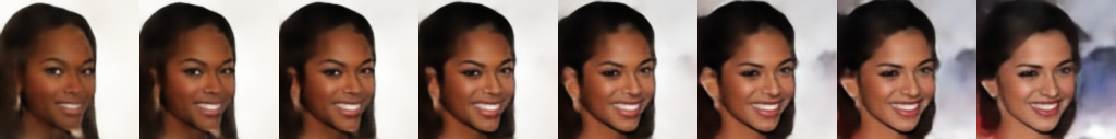} \\
  \includegraphics[width=0.99\linewidth]{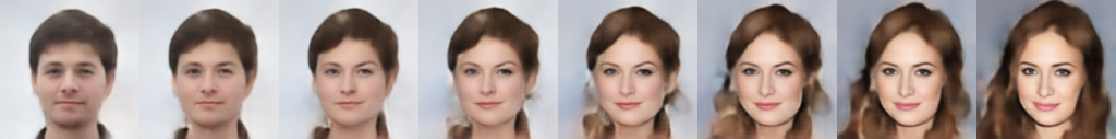} \\
  \includegraphics[width=0.99\linewidth]{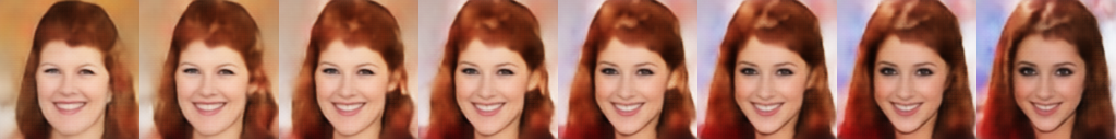} \\
  \includegraphics[width=0.99\linewidth]{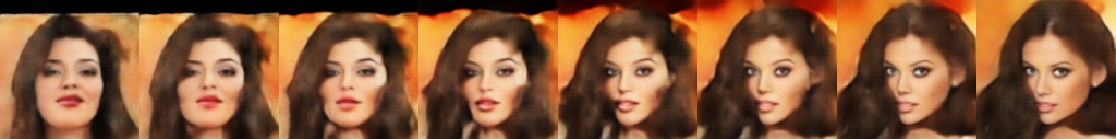} \\
  \includegraphics[width=0.99\linewidth]{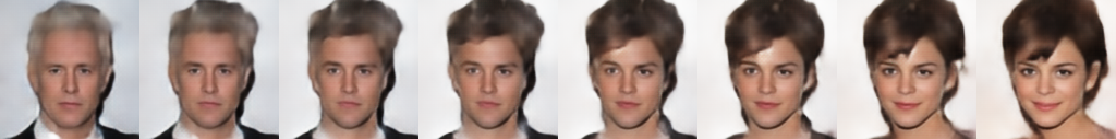}
  \caption{
    Illustration of interpolations obtained with our model on the CelebA dataset.
    Each row corresponds to an image pair, and the leftmost and rightmost images are actual images from the training set.
    Given two images $i$ and $j$, we get interpolated latent vectors $z$ between $z_i$ and $z_j$ and show the reconstruction $g(z)$.
  }
  \label{fig:path}
\end{figure}

%%%%%%%%%%%%%
%%% MODEL %%%
%%%%%%%%%%%%%

\section{The Generative Latent Optimization}\label{sec:gnat}
First, we consider a large set of images~$\{ x_1, \ldots, x_N \}$, where each
image~$x_i \in \mathcal{X}$ has dimensions~$3 \times w \times h$.  Second, we
initialize a set of~$d$-dimensional random vectors~$\{ z_1, \ldots, z_N \}$,
where $z_i \in \mathcal{Z} \subseteq \mathbb{R}^d$ for all~$i =1, \ldots N$.
Third, we pair the dataset of images with the random vectors, obtaining the
dataset~$\{(z_1, x_1), \ldots, (z_N, x_N) \}$.  Finally, we jointly learn the
parameters~$\theta$ in~$\Theta$ of a generator~$g_\theta : \mathcal{Z} \to
\mathcal{X}$ and the optimal noise vector~$z_i$ for each image~$x_i$, by
solving:
\begin{equation}
  \label{eq:nai}
  \min_{\theta\in\Theta}~~\frac{1}{N}\sum_{i=1}^N~\left[~\min_{z_i \in\mathcal{Z}}~~\ell\left(g_\theta(z_i),x_i\right)\right],
\end{equation}
In the previous,~$\ell : \mathcal{X} \times \mathcal{X}$ is a loss function
measuring the reconstruction error from~$g(z_i)$ to~$x_i$.  We call this model
Generative Latent Optimization~(\GNAT{}).

\noindent {\bf Learnable~$z_i$.} In contrast to autoencoders \citep{boulard_ae}, which assume a parametric model~$f :
\mathcal{X} \to \mathcal{Z}$, usually referred to as the \emph{encoder}, to compute
the vector~$z$ from samples~$x$, and minimize the reconstruction loss~$\ell(g(f(x)),x)$,  in \GNAT{} we jointly
optimize the inputs~$z_1, \ldots, z_N$ and the model parameter~$\theta$.
Since the vector~$z$ is a free
parameter, our model can recover all the solutions that could be found by an
autoencoder, and reach some others. In a nutshell,~\GNAT{} can be
viewed as an~``encoder-less'' autoencoder, or as a~``discriminator-less''~GAN.

\noindent {\bf Choice of~$\mathcal{Z}$.}
A common choice of $\mathcal{Z}$ in the GAN literature is from a Normal distribution
on~$\mathbb{R}^d$.  Since random vectors~$z$ drawn from the~$d$-dimensional
Normal distribution are very unlikely to land far outside the (surface of) the sphere~$\mathcal{S}(\sqrt{d}, d, 2)$, and since
projection onto the sphere is easy and numerically pleasant,  after each $z$ update in \GNAT{} training we project onto the sphere.
For simplicity, instead of using the $\sqrt{d}$ sphere, we use the unit sphere.

\noindent {\bf Choice of loss function.}
On the one hand, the squared-loss function~$\ell_2(x,x') = \| x - x' \|_2^2$ is
a simple choice, but leads to blurry (average) reconstructions of natural images.
On the other hand,~GANs use a~convnet~(the discriminator) as loss function. Since
the early layers of~convnets focus on edges, the samples from a~GAN are sharper.
Therefore, our experiments provide quantitative and qualitative comparisons
between the~$\ell_2$ loss and the Laplacian pyramid~$\text{Lap}_1$ loss
\begin{equation*}
    \label{eq:laploss}
	\text{Lap}_1(x,x') = \sum_{j} 2^{2j} |L^j(x) - L^j(x')|_1,
\end{equation*}
where~$L^j(x)$ is the~$j$-th level of the Laplacian pyramid representation of~$x$ \citep{lapl1cost}.
Therefore, the~$\text{Lap}_1$ loss weights the details
at fine scales more heavily.  In order to preserve low-frequency content such as color
information, we will use a weighted combination of the $\text{Lap}_1$ and the
$\ell_2$ costs.

\noindent {\bf Optimization.} For any choice of differentiable generator, the
objective \eqref{eq:nai} is differentiable with respect to~$z$, and~$\theta$.
Therefore, we will learn~$z$ and~$\theta$ by Stochastic Gradient Descent (SGD).
The gradient of \eqref{eq:nai} with respect to~$z$ can be obtained by
backpropagating the gradients through the generator
function~\citep{compressed}. We project each~$z$ back to the representation
space~$\mathcal{Z}$ after each update. To have noise vectors laying on the unit
$\ell_2$ sphere, we project $z$ after each update by dividing its value by
$\max(\|z\|_2, 1)$.
We initialize $z$ by sampling them from a Gaussian distribution.
\vspace{1em}

\noindent {\bf Generator architecture.}
Among the multiple architectural variations
explored in the literature, the most prominent is the Deep Convolutional
Generative Adversarial Network (DCGAN) \citep{dcgan}. Therefore, in this paper, to make the comparison with the GAN literature as straightforward as possible, we will use the generator function of DCGAN
construct the generator of \GNAT{} across all of our experiments.

%%%%%%%%%%%%%%%%%%%%
%%% RELATED WORK %%%
%%%%%%%%%%%%%%%%%%%%
\section{Related work}

\begin{figure}[t]
  \centering
  \includegraphics[width=0.99\linewidth]{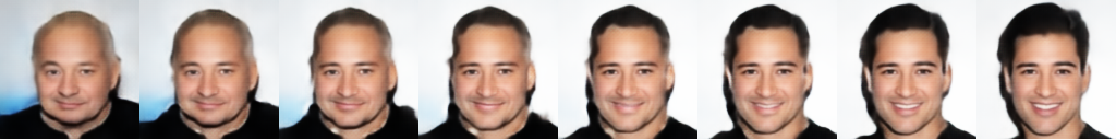} \\
  \includegraphics[width=0.99\linewidth]{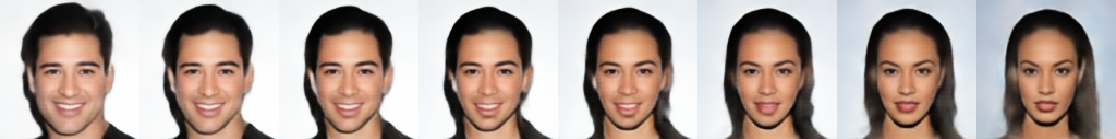} \\
  \includegraphics[width=0.99\linewidth]{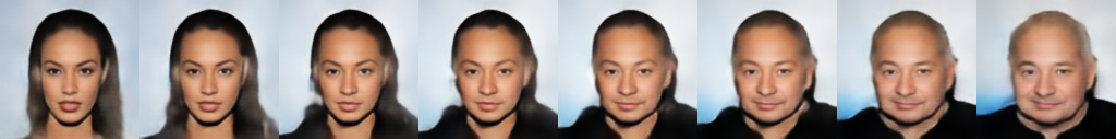}
  \caption{
    Illustration of interpolations obtained with our model on the CelebA dataset.
    We construct a path between $3$ images to verify that paths do not collapse to an ``average'' representation
    in the middle of the interpolation.
  }
  \label{fig:path2}
\end{figure}

\noindent {\bf Generative Adversarial Networks.}
GANs were introduced by \citet{gan}, and
refined in multiple recent
works~\citep{denton2015deep,dcgan,zhao2016energy,salimans2016improved}.
As described in Section~\ref{sec:introduction}, GANs construct a generative
model of a probability distribution $P$ by setting up an
adversarial game between a generator $g$ and a discriminator $d$:
\begin{equation*}
  \label{eq:gan}
  \min_G \max_D \mathbb{E}_{x\sim P}\, \log d(x) + \mathbb{E}_{z \sim Q}\,
  (1-\log d(g(z))).
\end{equation*}
In practice, most of the applications of GANs concern modeling
distributions of natural images. In these cases, both the generator $g$ and
the discriminator $d$ are parametrized as deep convnets
\citep{lecun1998gradient}.

\noindent {\bf Autoencoders.}
In their simplest form, an Auto-Encoder (AE) is a pair of neural networks, formed by
an encoder $f : \mathcal{X} \to \mathcal{Z}$ and a decoder $g : \mathcal{Z} \to
\mathcal{X}$. The role of an autoencoder is the compress the data $\{x_1,
\ldots, x_N\}$ into the representation $\{z_1, \ldots, z_N\}$ using the encoder
$f(x_i)$, and decompress it using the decoder $g(f(x_i))$. Therefore,
autoencoders minimize
$\mathbb{E}_{x \sim P}\,\ell(g(f(x)), x)$,
where $\ell : \mathcal{X} \times \mathcal{X}$ is a simple loss function, such
as the mean squared error.  There is a vast literature on autoencoders,
spanning three decades from their conception \citep{boulard_ae, baldi1989neural},
renaissance \citep{hinton_science}, and
recent
probabilistic extensions~\citep{vincent2008extracting,kingma2013auto}.

Several works have combined GANs with AEs. For instance, \citet{zhao2016energy}
replace the discriminator of a GAN by an AE, and \citet{adversarial_generator}
replace the decoder of an AE by a generator of a GAN.  Similar to \GNAT{},
these works suggest that the combination of standard pipelines can lead to good
generative models. In this work we attempt one step further, to
explore if learning a generator alone is possible.

\noindent {\bf Inverting generators.}
Several works attempt at recovering the latent representation of an image with
respect to a generator.  In particular, \citet{precise_recovery,
zhu2016generative} show that it is possible to recover $z$ from a {\it generated}
sample.  Similarly, \citet{inverting_generator} show that it is possible to
learn the inverse transformation of a generator.  These works are similar to
\citep{visualizing_cnn}, where the gradients of a particular feature of a
convnet are back-propagated to the pixel space in order to visualize what that
feature stands for. From a theoretical perspective, \citet{bruna2013signal}
explore the theoretical conditions for a network to be invertible.  All of
these inverting efforts are instances of the \emph{pre-image problem},
\citep{kwok2004pre}.

\citet{compressed} have recently showed that it is possible to recover from a
trained generator with compressed sensing. Similar to our work, they use a
$\ell_2$ loss and backpropagate the gradient to the low rank distribution.
However, they do not train the generator simultaneously. Jointly learning the
representation and training the generator allows us to extend their findings.
\citet{santurkar2017generative} also use generative models to compress images.

Several works have used an optimization of a latent representation for the
express purpose of generating realistic images, e.g.
\citep{,Portilla2000,nguyen2016ppgn}.  In these works, the total loss function
optimized to generate is trained separately from the optimization of the latent
representation (in the former, the loss is based on a complex wavelet
transform, and in the latter, on separately trained autoencoders and
classification convolutional networks). In this work we train the latent
representations and the generator together from scratch; and show that at test
time we may sample new $z$ either using simple parametric distributions or interpolations in the latent space.

\begin{figure}[t]
  \centering
  \includegraphics[width=0.49\linewidth]{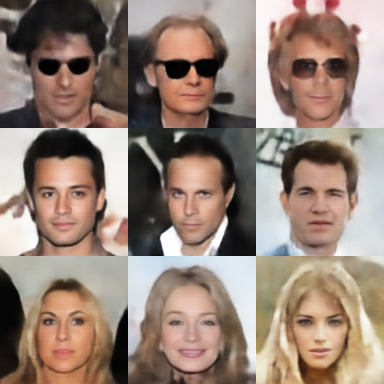} \
  \includegraphics[width=0.49\linewidth]{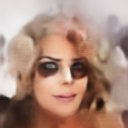}
  \caption{
    Illustration of feature arithmetic on the CelebA dataset.
    We show that by taking the average hidden representation of the first row (man with sunglasses), substracting the one of the second row (men without sunglasses) and adding the one of the third row (women without sunglasses), we obtain a coherent image.
  }
  \label{fig:arith}
\end{figure}

\noindent {\bf Learning representations.} Arguably, the problem of learning
representations from data in an unsupervised manner is one of the long-standing
problems in machine learning \citep{bengio2013representation, lecun2015deep}.
One of the earliest algorithms used to achieve is goal is Principal Component
Analysis, or PCA \citep{pca_original, pca}. For instance, PCA has been used to
learn low-dimensional representations of human faces \citep{eigenfaces}, or to
produce a hierarchy of features \citep{pcanet}. The nonlinear extension of PCA
is an autoencoder \citep{baldi1989neural}, which is in turn one of the most
extended algorithms to learn low-dimensional representations from data.
Similar algorithms learn low-dimensional representations of data with certain
structure. For instance, in sparse coding \citep{aharon2006rm,
mairal2008sparse}, the representation of one image is the linear combination of
a very few elements from a dictionary of features.  More recently,
\citet{understanding} realized the capability of deep neural networks to map
large collections of images to noise vectors, and \citet{nat} exploited a
similar procedure to learn visual features unsupervisedly.  Similarly to us,
\citet{nat} allow the noise vectors $z$ to move in order to better learn the
mapping from images to noise vectors. The proposed \GNAT{} is the analogous to
these works, in the opposite direction: learn a map \emph{from} noise vectors
\emph{to} images.  Finally, the idea of mapping between images and noise to
learn generative models is a well known technique \citep{gaussianization,
gauss_laparra, nonequilibrium, infusion}.

\noindent {\bf Nuisance Variables.} One might consider the generator parameters the
variables of interest, and $Z$ to be ``nuisance variables''. There is a classical
literature on dealing with nuisance parameters while estimating the parameters
of interest, including optimization methods as we have used~\citep{kendall}.
In this framing, it may be better to marginalize over the nuisance
variables, but for the models and data we use this is intractable.

\noindent {\bf Speech and music generation.}
Optimizing a latent representation of a generative model has a long history in speech~\cite{RabinerS07}, both for fitting single examples in the context of fitting a generative model, and in the context of speaker adaptation.
In the context of music generation and harmonazation, the first model was introduced by~\citet{ebciouglu1988expert}.
Closer to our work, is the neural network-based model of~\citet{hild1992harmonet}, which was later improved upon by~\citet{hadjeres2016deepbach}.

\begin{table*}[t]
  \centering
  \begin{tabular}{l c rr c rr c rr c rr c rr c rr}
    \toprule
    && \multicolumn{2}{c}{MNIST} && \multicolumn{2}{c}{SVHN}  && \multicolumn{5}{c}{CelebA} && \multicolumn{5}{c}{LSUN} \\
    && \multicolumn{2}{c}{32} && \multicolumn{2}{c}{32} && \multicolumn{2}{c}{64} && \multicolumn{2}{c}{128} && \multicolumn{2}{c}{64} && \multicolumn{2}{c}{128}  \\
    \cmidrule{3-4} \cmidrule{6-7} \cmidrule{9-10} \cmidrule{12-13} \cmidrule{15-16} \cmidrule{18-19}
    method && train & test  && train & test && train & test && train & test && train & test && train & test  \\
    \midrule
    PCA   && 20.6 & 20.3 && 30.2 & 30.3 && 25.1 & 25.1 && 23.6 & 23.6 && 23.6 & 23.7 && 21.9 & 22.0 \\
    \midrule
    VAE   && 26.2 & 25.7 && 27.9 & 27.8 && 25.0 & 24.9 && 26.2 & 25.0 && 23.8 & 23.8 && 22.1 & 22.1 \\
    DCGAN && 26.9 & 27.2 && 30.2 & 30.1 && 25.0 & 25.0 && 23.5 & 23.5 && 21.8 & 21.9 && 20.8 & 20.9 \\
    GLO   && 27.0 & 27.2 && 30.7 & 30.7 && 27.7 & 27.7 && 26.4 & 26.4 && 24.8 & 24.9 && 22.0 & 22.1 \\
    \midrule
    VAE   && 25.3 & 25.0 && 24.5 & 24.5 && 22.8 & 22.8 && 23.4 & 23.2 && 22.1 & 22.1 && 20.6 & 20.6 \\
    DCGAN && 25.8 & 26.2 && 26.0 & 26.0 && 21.9 & 21.9 && 21.3 & 21.3 && 19.0 & 19.1 && 18.7 & 18.7 \\
    GLO   && 26.2 & 26.2 && 27.9 & 28.0 && 25.5 & 25.6 && 24.7 & 24.8 && 23.3 & 23.4 && 21.4 & 21.4 \\
    \bottomrule
  \end{tabular}
  \caption{
    pSNR of reconstruction for different models.  Below the line, the codes were found using  $\text{Lap}_1$ loss (although the test error is still measured in pSNR).   Above the line, the codes were found using mean square error.  Note that the generators of the VAE and \GNAT{} models were trained to reconstruct in   $\text{Lap}_1$  loss.   pSNR of GAN reconstruction of images generated by GAN (not real images) is greater than 50. }
  \label{tab:psnr}
  \vspace{-0.5em}
\end{table*}

\begin{figure*}[t]
  \centering
	\begin{subfigure}[b]{0.49\linewidth}
    \includegraphics[width=0.03 \textwidth]{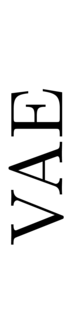}
    \includegraphics[width=0.96\textwidth]{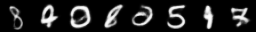} \\
    \includegraphics[width=0.03 \textwidth]{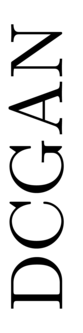}
    \includegraphics[width=0.96\textwidth]{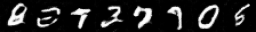} \\
    \includegraphics[width=0.03 \textwidth]{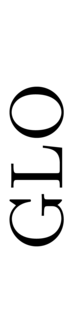}
    \includegraphics[width=0.96\textwidth]{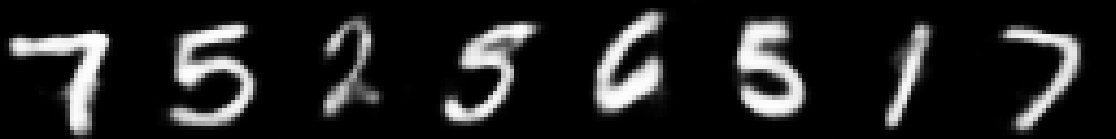}
    \caption{MNIST}
  \end{subfigure}
	\
	\begin{subfigure}[b]{0.49\linewidth}
    \includegraphics[width=0.03 \textwidth]{figures-vae-01-small.png}
    \includegraphics[width=0.96\textwidth]{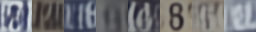} \\
    \includegraphics[width=0.03 \textwidth]{figures-dcgan-01-small.png}
    \includegraphics[width=0.96\textwidth]{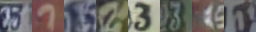} \\
    \includegraphics[width=0.03 \textwidth]{figures-glo-01-small.png}
    \includegraphics[width=0.96\textwidth]{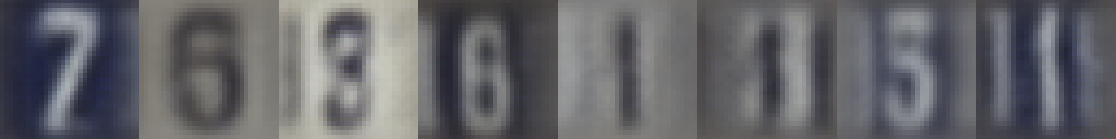}
    \caption{SVHN}
  \end{subfigure}
	\\
  \vspace{0.5em}
	\begin{subfigure}[b]{0.49\textwidth}
    \includegraphics[width=0.03 \textwidth]{figures-vae-01-small.png}
    \includegraphics[width=0.96\textwidth]{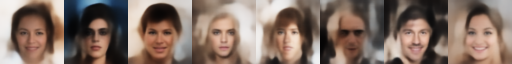} \\
    \includegraphics[width=0.03 \textwidth]{figures-dcgan-01-small.png}
    \includegraphics[width=0.96\textwidth]{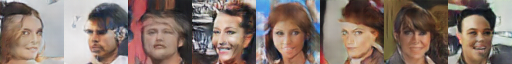} \\
    \includegraphics[width=0.03 \textwidth]{figures-glo-01-small.png}
    \includegraphics[width=0.96\textwidth]{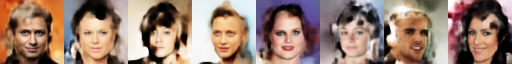}
    \caption{CelebA-64}
  \end{subfigure}
	\
	\begin{subfigure}[b]{0.49\textwidth}
    \includegraphics[width=0.03 \textwidth]{figures-vae-01-small.png}
    \includegraphics[width=0.96\textwidth]{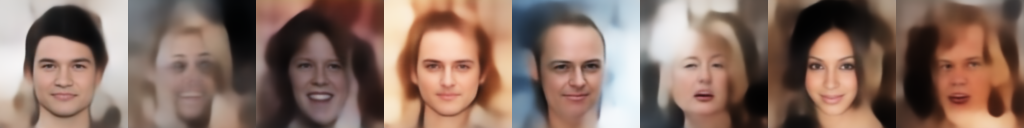} \\
    \includegraphics[width=0.03 \textwidth]{figures-dcgan-01-small.png}
    \includegraphics[width=0.96\textwidth]{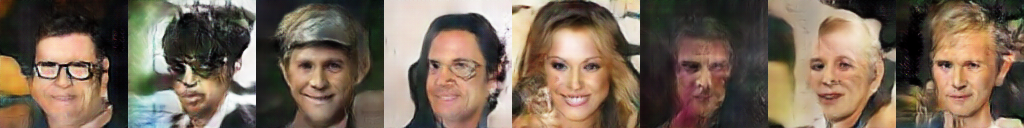} \\
    \includegraphics[width=0.03 \textwidth]{figures-glo-01-small.png}
    \includegraphics[width=0.96\textwidth]{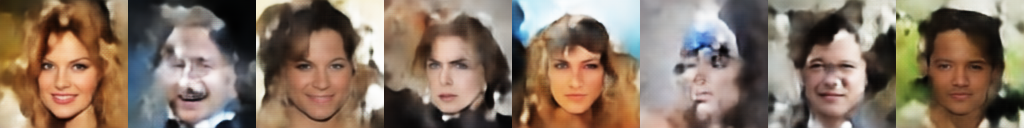}
    \caption{CelebA-128}
  \end{subfigure}
	\\
  \vspace{0.5em}
	\begin{subfigure}[b]{0.49\textwidth}
    \includegraphics[width=0.03 \textwidth]{figures-vae-01-small.png}
    \includegraphics[width=0.96\textwidth]{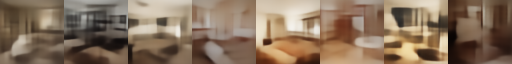} \\
    \includegraphics[width=0.03 \textwidth]{figures-dcgan-01-small.png}
    \includegraphics[width=0.96\textwidth]{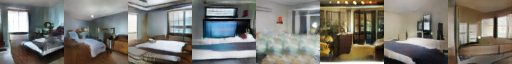} \\
    \includegraphics[width=0.03 \textwidth]{figures-glo-01-small.png}
    \includegraphics[width=0.96\textwidth]{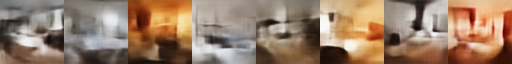}
    \caption{LSUN-64}
  \end{subfigure}
  \
	\begin{subfigure}[b]{0.49\textwidth}
    \includegraphics[width=0.03 \textwidth]{figures-vae-01-small.png}
    \includegraphics[width=0.96\textwidth]{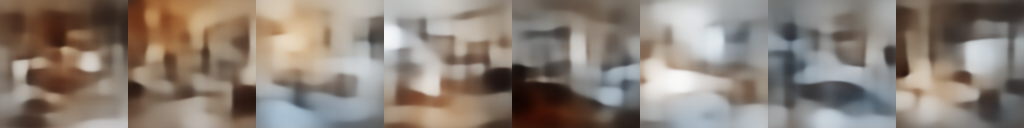} \\
    \includegraphics[width=0.03 \textwidth]{figures-dcgan-01-small.png}
    \includegraphics[width=0.96\textwidth]{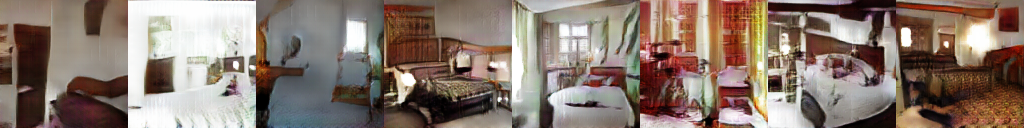} \\
    \includegraphics[width=0.03 \textwidth]{figures-glo-01-small.png}
    \includegraphics[width=0.96\textwidth]{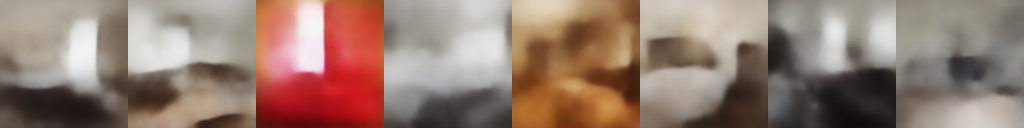}
    \caption{LSUN-128}
  \end{subfigure}
  \caption{Samples generated by VAE, DCGAN and \GNAT{} on the $4$ datasets. For CelebA and LSUN, we consider images of size $64$ and $128$.
  On small datasets, the three models generate images of the comparable quality. On LSUN, images from VAE and \GNAT{} are nowhere close to those from DCGAN.}
  \label{fig:samples}
  \vspace{-0.5em}
\end{figure*}

\begin{figure*}[h!]
  \centering
	\begin{subfigure}[b]{0.49\linewidth}
    \includegraphics[width=0.03 \textwidth]{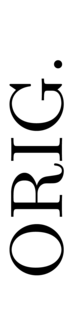}
    \includegraphics[width=0.96\textwidth]{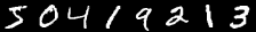}
    \includegraphics[width=0.03 \textwidth]{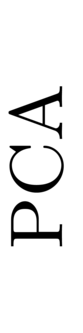}
    \includegraphics[width=0.96\textwidth]{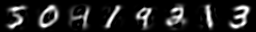}
    \includegraphics[width=0.03 \textwidth]{figures-vae-01-small.png}
    \includegraphics[width=0.96\textwidth]{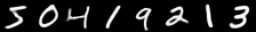}
    \includegraphics[width=0.03 \textwidth]{figures-dcgan-01-small.png}
    \includegraphics[width=0.96\textwidth]{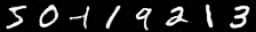}
    \includegraphics[width=0.03 \textwidth]{figures-glo-01-small.png}
    \includegraphics[width=0.96\textwidth]{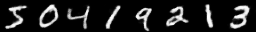}
    \caption{MNIST}
  \end{subfigure}
	\
	\begin{subfigure}[b]{0.49\linewidth}
    \includegraphics[width=0.03 \textwidth]{figures-orig-01-small.png}
    \includegraphics[width=0.96\textwidth]{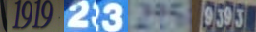}
    \includegraphics[width=0.03 \textwidth]{figures-pca-01-small.png}
    \includegraphics[width=0.96\textwidth]{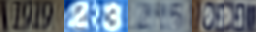}
    \includegraphics[width=0.03 \textwidth]{figures-vae-01-small.png}
    \includegraphics[width=0.96\textwidth]{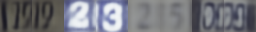}
    \includegraphics[width=0.03 \textwidth]{figures-dcgan-01-small.png}
    \includegraphics[width=0.96\textwidth]{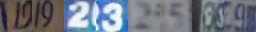}
    \includegraphics[width=0.03 \textwidth]{figures-glo-01-small.png}
    \includegraphics[width=0.96\textwidth]{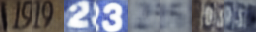}
    \caption{SVHN}
  \end{subfigure}
	\\
  \vspace{0.5em}
	\begin{subfigure}[b]{0.49\textwidth}
    \includegraphics[width=0.03 \textwidth]{figures-orig-01-small.png}
    \includegraphics[width=0.96\textwidth]{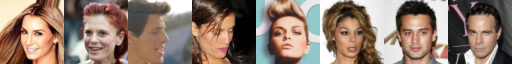}
    \includegraphics[width=0.03 \textwidth]{figures-pca-01-small.png}
    \includegraphics[width=0.96\textwidth]{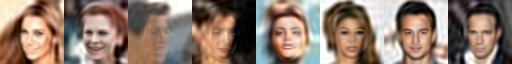}
    \includegraphics[width=0.03 \textwidth]{figures-vae-01-small.png}
    \includegraphics[width=0.96\textwidth]{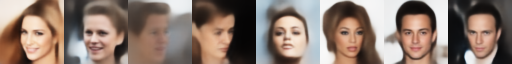}
    \includegraphics[width=0.03 \textwidth]{figures-dcgan-01-small.png}
    \includegraphics[width=0.96\textwidth]{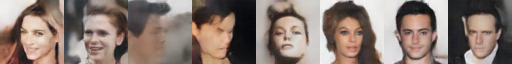}
    \includegraphics[width=0.03 \textwidth]{figures-glo-01-small.png}
    \includegraphics[width=0.96\textwidth]{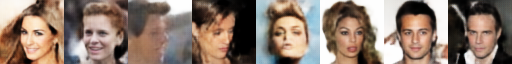}
    \caption{CelebA-64}
  \end{subfigure}
	\
	\begin{subfigure}[b]{0.49\textwidth}
    \includegraphics[width=0.03 \textwidth]{figures-orig-01-small.png}
    \includegraphics[width=0.96\textwidth]{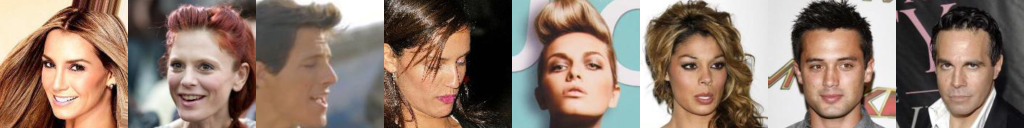}
    \includegraphics[width=0.03 \textwidth]{figures-pca-01-small.png}
    \includegraphics[width=0.96\textwidth]{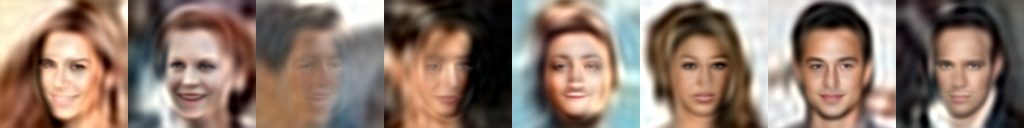}
    \includegraphics[width=0.03 \textwidth]{figures-vae-01-small.png}
    \includegraphics[width=0.96\textwidth]{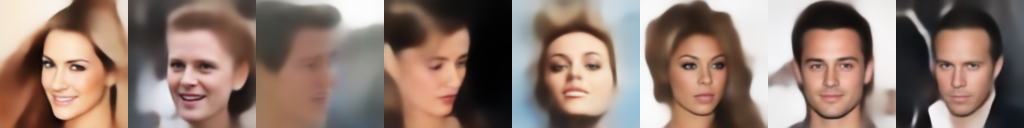}
    \includegraphics[width=0.03 \textwidth]{figures-dcgan-01-small.png}
    \includegraphics[width=0.96\textwidth]{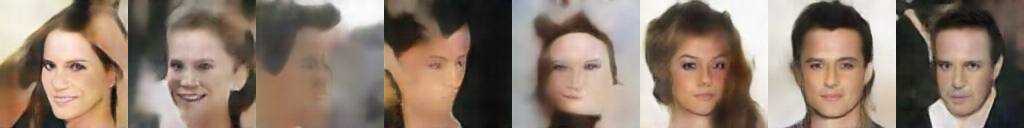}
    \includegraphics[width=0.03 \textwidth]{figures-glo-01-small.png}
    \includegraphics[width=0.96\textwidth]{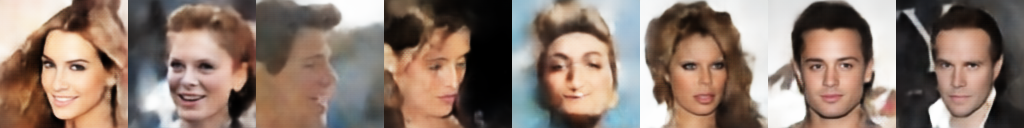}
    \caption{CelebA-128}
  \end{subfigure}
	\\
  \vspace{0.5em}
	\begin{subfigure}[b]{0.49\textwidth}
    \includegraphics[width=0.03 \textwidth]{figures-orig-01-small.png}
    \includegraphics[width=0.96\textwidth]{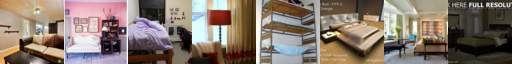}
    \includegraphics[width=0.03 \textwidth]{figures-pca-01-small.png}
    \includegraphics[width=0.96\textwidth]{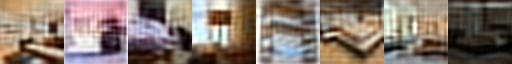}
    \includegraphics[width=0.03 \textwidth]{figures-vae-01-small.png}
    \includegraphics[width=0.96\textwidth]{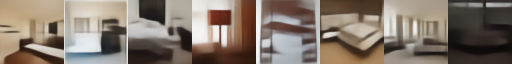}
    \includegraphics[width=0.03 \textwidth]{figures-dcgan-01-small.png}
    \includegraphics[width=0.96\textwidth]{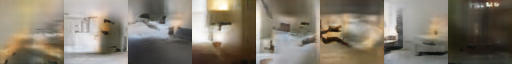}
    \includegraphics[width=0.03 \textwidth]{figures-glo-01-small.png}
    \includegraphics[width=0.96\textwidth]{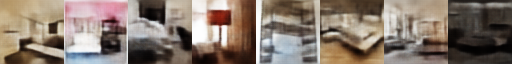}
    \caption{LSUN-64}
  \end{subfigure}
  \
	\begin{subfigure}[b]{0.49\textwidth}
    \includegraphics[width=0.03 \textwidth]{figures-orig-01-small.png}
    \includegraphics[width=0.96\textwidth]{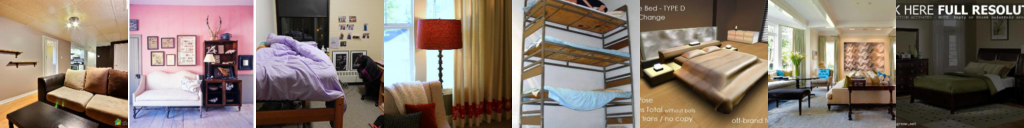}
    \includegraphics[width=0.03 \textwidth]{figures-pca-01-small.png}
    \includegraphics[width=0.96\textwidth]{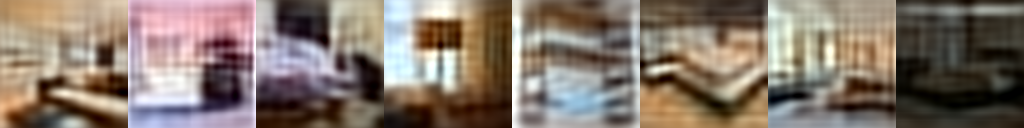}
    \includegraphics[width=0.03 \textwidth]{figures-vae-01-small.png}
    \includegraphics[width=0.96\textwidth]{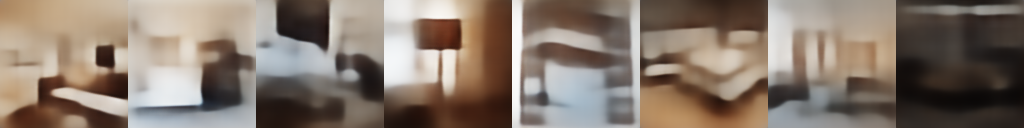}
    \includegraphics[width=0.03 \textwidth]{figures-dcgan-01-small.png}
    \includegraphics[width=0.96\textwidth]{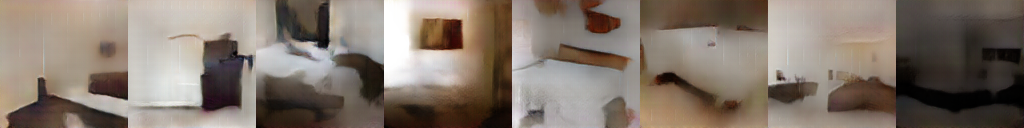}
    \includegraphics[width=0.03 \textwidth]{figures-glo-01-small.png}
    \includegraphics[width=0.96\textwidth]{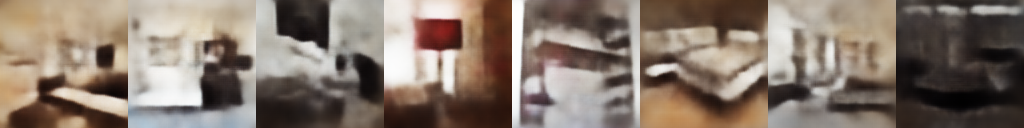}
    \caption{LSUN-128}
  \end{subfigure}
  \caption{Reconstruction results from PCA, VAE, DCGAN and \GNAT{} on the $4$ datasets. The original images are on the top row.
  VAE reconstructions are blurrier than \GNAT{} .  DCGAN fails to reconstruct images from large datasets.}
  \label{fig:reconstructions}
  \vspace{-0.5em}
\end{figure*}

\section{Experiments}
\label{sec:experiments}

In this section, we compare \GNAT{} quantitatively and qualitatively against standard generative models on a variety of datasets.
We consider several tasks to understand the strengths and weaknesses of each model:
a qualitative analysis of the properties of the latent space typically observed with deep generative models and
an image reconstruction problem to give some quantitative insights on the capability of \GNAT{} to cover a dataset.
We selected datasets that are both small and large, uni-modal and multi-modal to stress the specificities of our models in different settings.

\begin{figure}
  1\textsuperscript{st} principal vector \\
  \includegraphics[width=0.99\linewidth]{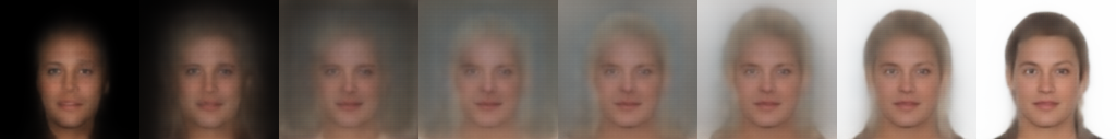} \\
  2\textsuperscript{nd} principal vector \\
  \includegraphics[width=0.99\linewidth]{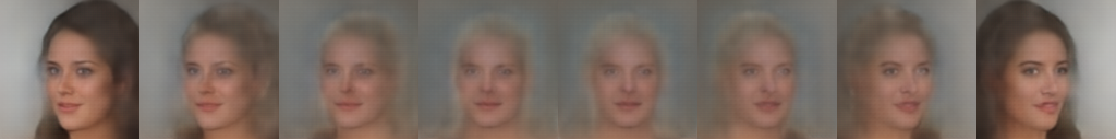} \\
  3\textsuperscript{rd} principal vector \\
  \includegraphics[width=0.99\linewidth]{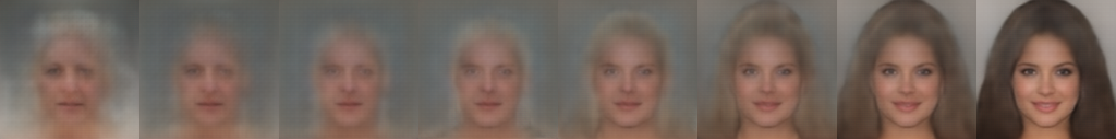} \\
  4\textsuperscript{th} principal vector \\
  \includegraphics[width=0.99\linewidth]{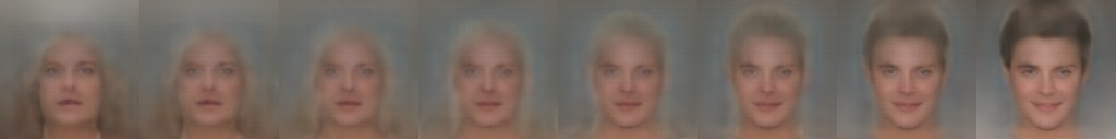}
  \caption{
  	Interpolation from the average face along the principal vectors corresponding to the largest principal values.  The principal vectors
  	and values were computed on the (centered) $Z$ vectors of the GLO model.
    The first vector seems to capture the brightness of the background, the second one the orientation of the face,
    while the third and forth capture the information about the gender.  The average image is 4th from left.
  }
  \label{fig:prop}
\end{figure}

\paragraph{Implementation details}
The generator of a \GNAT{} follows the same architecture as the generator of DCGAN.
We use Stochastic Gradient Descent (SGD) to optimize both $\theta$ and $z$, setting
the learning rate for $\theta$ at $1$ and the learning rate of $z$ at $10$.
After each update, the noise vectors $z$ are projected to the unit $\ell_2$
Sphere. In the sequel, we initialize the random vectors of GLO using a Gaussian
distribution (for the CelebA dataset) or the top $d$ principal components (for
the LSUN dataset).
We use the $\ell_2+\text{Lap}_1$ loss for all the experiments but MNIST where we use an MSE loss.

\subsection{Baselines and datasets.}
We consider three standard baselines: PCA, VAE, and GAN.
PCA~\citep{pca_original} is equivalent to a linear autoencoder \citep{baldi1989neural}.
We use for VAE and GAN the same generator architecture as for GLO, i.e., a DCGAN.
We also set the number of principal components for PCA to be same as the dimensions of the latent spaces.
We use 32 dimensions for MNIST, 64 dimensions for SVHN and 256 dimensions for CelebA and LSUN.
We use the same $\ell_2+\text{Lap}_1$ loss for VAE as for \GNAT{} for all the experiments but MNIST where we use an MSE loss.
For the rest, we train VAE with the default hyper-parameters for $25$ epochs.
We train the GAN baseline with the default hyper-parameters and many seeds.

For our empirical evaluation, we consider four varied image datasets.  We
select both ``unimodal'' and ``multimodal'' datasets to probe the difficulty of
models in each setting.  We carry out our experiments on
MNIST~\footnote{\url{http://yann.lecun.com/exdb/mnist/}},
SVHN\footnote{\url{http://ufldl.stanford.edu/housenumbers/}} as well as more
challenging datasets such as
CelebA\footnote{\url{http://mmlab.ie.cuhk.edu.hk/projects/CelebA.html}} and
LSUN-bedroom\footnote{\url{http://lsun.cs.princeton.edu/2017/}}.  On smaller
datasets (MNIST and SVHN), we keep the images $32$ pixels large.  For CelebA
and LSUN we resize the images to either $64$ and $128$ pixels large.  For each
dataset, we set aside evenly-spaced images corresponding to $\frac{1}{32}$ of
the data, and consider these images as a test set.  We train our models on the complement.

\subsection{Properties of the latent space}
The latent space of GANs seems to linearize the space of images. That is:
interpolations between a pair of $z$ vectors in the latent space map through
the generator to a semantically meaningful, smooth nonlinear interpolation in
image space.
Figure~\ref{fig:path} shows that the latent space of \GNAT{} seems to
linearize the image space as well. For
example, the model interpolates between examples that are geometrically quite
different, reconstructing the rotation of the head from left to right, as well
as interpolating between genders or different ages.
It is important to note that these paths do not go through an ``average'' image
of the dataset as the path interpolation between the $3$ images of Figure~\ref{fig:path2} shows.

Linear arithmetic operations in the latent space of GANs can lead to meaningful image transformations.
For example: (man with sunglasses - man +
woman) produces an image of a woman with sunglasses.
Figure~\ref{fig:arith} shows that the latent space of \GNAT{} shares the same property.

Finally, \GNAT{} models have the attractive property that the principal vectors corresponding to
the largest principal values are meaningful in image space. As shown in Figure~\ref{fig:prop}, they carry
information like background color, the orientation of the head and gender. Interestingly, the gender
information is represented by two principal vectors, one for the female and one for the male.

These results suggest that the desirable linearization properties of generators are probably due to the structure of the model (convnets) rather than the training procedure.

\subsection{Generation}
Another celebrated aspect of GANs is the high quality of the examples they generate.   To sample from a \GNAT{} model, we fit a single full-covariance Gaussian to the $Z$ found by the training procedure; and then pass samples from that Gaussian through the generator.
Figure~\ref{fig:samples} shows a comparison between images generated by VAE, GAN and \GNAT{} models trained on different datasets, offering a few insights on the main difference between the methods:
First, the images produced by VAE are often less sharp than GLO, in particular on large datasets like CelebA and LSUN bedroom.
This observation suggests that the prior distribution on the latent space of a VAE may be too strong to fit many images, while vectors in the latent space of \GNAT{} move freely and use as much space as required to fit the images in the latent space.   
On the other hand, on these datasets,  the trained $Z$ from \GNAT{} are Gaussian enough to produce decent generations when fit with a single (full-covariance) Gaussian.

Second, it is interesting to notice that on the LSUN bedrooms, VAE and GLO are much worse than GAN.
While they seem to capture the general shape of the bedrooms, they fail to produce the same level of detail as is observed in the samples generated by a GAN.
One possibility is that in these settings, the  ``mode dropping'' problem commonly discussed in the GAN literature ~\cite{gan_tutorial}  is more a feature than a bug.
Both VAE and GLO do not suffer from mode dropping by construction (since their loss forces them to reconstruct the whole dataset) and it is possible that as a result, they both generate poorly when the variability in the distribution increases relative to the model capacity.  In other words, when confronted with more data variability than it can handle, a GAN can still be successful in generating (and well-organizing) a well-chosen subset of the data.

In the next section, we look at the reconstruction error of each method on the different datasets. This quantitative evaluation gives further insights on the differences between the approaches, and in particular, it gives evidence that GANs are not covering the training data.\footnote{Here ``reduced'' or "not covering" may be in the sense of missing some examples, for example dropping a cluster from a Gaussian mixture model, or more subtle retreats from the full data, for example projecting onto some complicated sub-manifold.  We believe understanding precisely what reduction happens (if any) in the case of GANs trained with convnets on images is an exciting direction for future work.}

\subsection{Image reconstruction}
In this set of experiments, we evaluate the quality of image reconstructions for each method.
In Table~\ref{tab:psnr} we report the reconstruction error in pSNR, which for a given image I and a reconstruction R, is defined as:
\begin{equation}
\text{pSNR}(I, R) = - 20 \log_{10} \frac{\text{MAX}(I)}{\sqrt{\text{MSE}(I,R)}},
\end{equation}
where MAX corresponds to the maximal value the image I can attain, and MSE is the Mean Squared Error.

To reconstruct an image from the test set, we need to find its latent representation.   For the PCA and VAE baselines, this is straightforward.
The latent codes for DCGAN and \GNAT{} can be found by backpropagating the reconstruction error to the code through the generator.
Note that the generating functions of {\it all}  the \GNAT{} and VAE models in the table were trained with $\text{Lap}_1$ cost.
This discrepancy in training loss favors PCA and if we find codes using MSE for \GNAT{} and DCGAN instead of $\text{Lap}_1$, the scores improve by $1-2$ points, even though we did not train \GNAT{} with an MSE. 

Measuring a reconstruction error favors VAE and \GNAT{} over DCGAN as they are trained to minimize such an error metric. 
However, it is interesting to notice that on small datasets, there is no clear difference with DCGAN.
The difference in performance between DCGAN and the other methods increases with the size of the dataset.
This result already suggests that as the dataset grows, GANs are probably focusing on a subset of it, while, by objective, VAE and GLO are forced to reconstruct the full dataset. 
It is not clear though what is the nature of this ``subset'', as the distribution of the pSNR scores of a DCGAN is not significantly different from those of VAE or \GNAT{} as shown in the supplementary material.   Finally, we remark although it is a-priori possible that the process of finding codes via backpropagation is not succeeding with the GAN generators, we find in practice that when we reconstruct an image {\it generated by the GAN}, the results are nearly perfect ($\text{pSNR} > 50$).  This suggests that the difference in pSNR between the models is not due to poor optimization of the codes.

Figure~\ref{fig:reconstructions} shows qualitative examples of reconstruction.
As suggested by the quantitative results, the VAE reconstruction is much blurrier than \GNAT{} and the reconstruction quality of DCGAN quickly deteriorates with the size and variability of the dataset.
More interestingly, on CelebA, we observe that, while DCGAN reconstructions of frontal faces look good, DCGAN struggles on side faces as well as rare examples, e.g., stylistic or blurry images.
More important, they seem to be copy-pasting ``faces'' rather than reconstructing them.
This effect is even more apparent on LSUN where it is almost impossible to find an entire well reconstructed image.
However, even though the colors are off, the edges are sharp if they are reconstructed, suggesting that GANs are indeed focusing on some specificities of the image distribution.

%%%%%%%%%%%%%%%%%%
%%% DISCUSSION %%%
%%%%%%%%%%%%%%%%%%

\section{Discussion}\label{sec:discussion}
The experimental results presented in this work suggest that, when working with images, we can recover many of the properties of GANs using convnets trained with a simple reconstruction losses. 
While this does not invalidate the promise of GANs as generic models of uncertainty or as methods for building generative models, our results suggest that, in order to further test the adversarial construction, research needs to move beyond images modeled using convnets.
On the other hand, practitioners who care only about generating images for a particular application, and find that the parameterized discriminator does improve their results, can incorporate reconstruction losses in their models, alleviating some of the instability of adversarial training.

While the visual quality of our results are promising, especially on the CelebA dataset, they are not yet to the level of the results obtained by GANs on the LSUN bedrooms.
This suggests that being able to cover the entire dataset is too onerous of a task if all that is required is to generate a few nice samples.
In that respect, we see that GANs have trouble reconstructing randomly chosen images at the same level of fidelity as their generations.
At the same time, GANs can produce good images after a single pass through the data with SGD, suggesting that the so-called ``mode dropping'' can be seen as a feature.
In future work we hope to better understand the tension between these two observations, and clarify the definition of this phenomenon.

There are many possibilities for improving the quality of GLO samples beyond understanding the effects of coverage.
For example other loss functions (e.g. a VGG metric, as in \cite{nguyen2016ppgn}), model architectures, especially progressive generation \citep{karras2017progressive}, and more sophisticated sampling methods after training the model all may improve the visual quality GLO samples.
Finally, because the methods keep track of the correspondence between samples and their representatives, we hope to be able to organize the $Z$ in interesting ways as we train.

\bibliographystyle{icml2018}
\bibliography{paper}

\begin{thebibliography}{49}
\providecommand{\natexlab}[1]{#1}
\providecommand{\url}[1]{\texttt{#1}}
\expandafter\ifx\csname urlstyle\endcsname\relax
  \providecommand{\doi}[1]{doi: #1}\else
  \providecommand{\doi}{doi: \begingroup \urlstyle{rm}\Url}\fi

\bibitem[Aharon et~al.(2006)Aharon, Elad, and Bruckstein]{aharon2006rm}
Aharon, M., Elad, M., and Bruckstein, A.
\newblock $ rm k $-svd: An algorithm for designing overcomplete dictionaries
  for sparse representation.
\newblock \emph{IEEE Transactions on signal processing}, 2006.

\bibitem[Baldi \& Hornik(1989)Baldi and Hornik]{baldi1989neural}
Baldi, P. and Hornik, K.
\newblock Neural networks and principal component analysis: Learning from
  examples without local minima.
\newblock \emph{Neural networks}, 1989.

\bibitem[Bengio et~al.(2013)Bengio, Courville, and
  Vincent]{bengio2013representation}
Bengio, Y., Courville, A., and Vincent, P.
\newblock Representation learning: A review and new perspectives.
\newblock \emph{IEEE transactions on pattern analysis and machine
  intelligence}, 35\penalty0 (8):\penalty0 1798--1828, 2013.

\bibitem[{Bojanowski} \& {Joulin}(2017){Bojanowski} and {Joulin}]{nat}
{Bojanowski}, P. and {Joulin}, A.
\newblock {Unsupervised Learning by Predicting Noise}.
\newblock In \emph{ICML}, 2017.

\bibitem[{Bora} et~al.(2017){Bora}, {Jalal}, {Price}, and
  {Dimakis}]{compressed}
{Bora}, A., {Jalal}, A., {Price}, E., and {Dimakis}, A.~G.
\newblock {Compressed Sensing using Generative Models}.
\newblock \emph{arXiv preprint arXiv:1703.03208}, 2017.

\bibitem[{Bordes} et~al.(2017){Bordes}, {Honari}, and {Vincent}]{infusion}
{Bordes}, F., {Honari}, S., and {Vincent}, P.
\newblock {Learning to Generate Samples from Noise through Infusion Training}.
\newblock \emph{arXiv preprint arXiv:1703.06975}, 2017.

\bibitem[Bourlard \& Kamp(1988)Bourlard and Kamp]{boulard_ae}
Bourlard, H. and Kamp, Y.
\newblock Auto-association by multilayer perceptrons and singular value
  decomposition.
\newblock \emph{Biological cybernetics}, 1988.

\bibitem[Bruna et~al.(2013)Bruna, Szlam, and LeCun]{bruna2013signal}
Bruna, J., Szlam, A., and LeCun, Y.
\newblock Signal recovery from pooling representations.
\newblock \emph{arXiv preprint arXiv:1311.4025}, 2013.

\bibitem[{Chan} et~al.(2015){Chan}, {Jia}, {Gao}, {Lu}, {Zeng}, and
  {Ma}]{pcanet}
{Chan}, T.-H., {Jia}, K., {Gao}, S., {Lu}, J., {Zeng}, Z., and {Ma}, Y.
\newblock {PCANet: A Simple Deep Learning Baseline for Image Classification?}
\newblock \emph{IEEE Transactions on Image Processing}, 2015.

\bibitem[Chen \& Gopinath(2000)Chen and Gopinath]{gaussianization}
Chen, S.~S. and Gopinath, R.~A.
\newblock Gaussianization.
\newblock In \emph{NIPS}, 2000.

\bibitem[{Creswell} \& {Bharath}(2016){Creswell} and
  {Bharath}]{inverting_generator}
{Creswell}, A. and {Bharath}, A.~A.
\newblock {Inverting The Generator Of A Generative Adversarial Network}.
\newblock \emph{arXiv preprints arXiv:1611.05644}, 2016.

\bibitem[Denton et~al.(2015)Denton, Chintala, Szlam, and
  Fergus]{denton2015deep}
Denton, E.~L., Chintala, S., Szlam, A., and Fergus, R.
\newblock Deep generative image models using a laplacian pyramid of adversarial
  networks.
\newblock In \emph{NIPS}, 2015.

\bibitem[Donahue et~al.(2016)Donahue, Kr{\"a}henb{\"u}hl, and
  Darrell]{donahue2016adversarial}
Donahue, J., Kr{\"a}henb{\"u}hl, P., and Darrell, T.
\newblock Adversarial feature learning.
\newblock \emph{arXiv preprint arXiv:1605.09782}, 2016.

\bibitem[Dumoulin et~al.(2016)Dumoulin, Belghazi, Poole, Lamb, Arjovsky,
  Mastropietro, and Courville]{dumoulin2016adversarially}
Dumoulin, V., Belghazi, I., Poole, B., Lamb, A., Arjovsky, M., Mastropietro,
  O., and Courville, A.
\newblock Adversarially learned inference.
\newblock \emph{arXiv preprint arXiv:1606.00704}, 2016.

\bibitem[Ebcio{\u{g}}lu(1988)]{ebciouglu1988expert}
Ebcio{\u{g}}lu, K.
\newblock An expert system for harmonizing four-part chorales.
\newblock \emph{Computer Music Journal}, 12\penalty0 (3):\penalty0 43--51,
  1988.

\bibitem[{Goodfellow}(2017)]{gan_tutorial}
{Goodfellow}, I.
\newblock {NIPS 2016 Tutorial: Generative Adversarial Networks}.
\newblock \emph{arXiv preprint arXiv:1701.00160}, 2017.

\bibitem[Goodfellow et~al.(2014)Goodfellow, Pouget-Abadie, Mirza, Xu,
  Warde-Farley, Ozair, Courville, and Bengio]{gan}
Goodfellow, I., Pouget-Abadie, J., Mirza, M., Xu, B., Warde-Farley, D., Ozair,
  S., Courville, A., and Bengio, Y.
\newblock Generative adversarial nets.
\newblock In \emph{NIPS}, 2014.

\bibitem[Hadjeres \& Pachet(2017)Hadjeres and Pachet]{hadjeres2016deepbach}
Hadjeres, G. and Pachet, F.
\newblock Deepbach: a steerable model for bach chorales generation.
\newblock \emph{ICML}, 2017.

\bibitem[Hild et~al.(1992)Hild, Feulner, and Menzel]{hild1992harmonet}
Hild, H., Feulner, J., and Menzel, W.
\newblock Harmonet: A neural net for harmonizing chorales in the style of js
  bach.
\newblock In \emph{NIPS}, 1992.

\bibitem[Hinton \& Salakhutdinov(2006)Hinton and Salakhutdinov]{hinton_science}
Hinton, G.~E. and Salakhutdinov, R.~R.
\newblock Reducing the dimensionality of data with neural networks.
\newblock \emph{Science}, 2006.

\bibitem[Iizuka et~al.(2017)Iizuka, Simo-Serra, and
  Ishikawa]{IizukaSIGGRAPH2017}
Iizuka, S., Simo-Serra, E., and Ishikawa, H.
\newblock {Globally and Locally Consistent Image Completion}.
\newblock \emph{ACM Transactions on Graphics}, 36\penalty0 (4):\penalty0
  107:1--107:14, 2017.

\bibitem[Jolliffe(1986)]{pca}
Jolliffe, I.~T.
\newblock \emph{Principal component analysis and factor analysis}.
\newblock Springer, 1986.

\bibitem[Karras et~al.(2017)Karras, Aila, Laine, and
  Lehtinen]{karras2017progressive}
Karras, T., Aila, T., Laine, S., and Lehtinen, J.
\newblock Progressive growing of gans for improved quality, stability, and
  variation.
\newblock \emph{arXiv preprint arXiv:1710.10196}, 2017.

\bibitem[Kingma \& Welling(2013)Kingma and Welling]{kingma2013auto}
Kingma, D.~P. and Welling, M.
\newblock Auto-encoding variational bayes.
\newblock \emph{arXiv preprint arXiv:1312.6114}, 2013.

\bibitem[Kwok \& Tsang(2004)Kwok and Tsang]{kwok2004pre}
Kwok, J.-Y. and Tsang, I.-H.
\newblock The pre-image problem in kernel methods.
\newblock \emph{IEEE transactions on neural networks}, 2004.

\bibitem[Laparra et~al.(2011)Laparra, Camps-Valls, and Malo]{gauss_laparra}
Laparra, V., Camps-Valls, G., and Malo, J.
\newblock Iterative gaussianization: from ica to random rotations.
\newblock \emph{IEEE transactions on neural networks}, 2011.

\bibitem[LeCun et~al.(1998)LeCun, Bottou, Bengio, and
  Haffner]{lecun1998gradient}
LeCun, Y., Bottou, L., Bengio, Y., and Haffner, P.
\newblock Gradient-based learning applied to document recognition.
\newblock \emph{Proceedings of the IEEE}, 1998.

\bibitem[LeCun et~al.(2015)LeCun, Bengio, and Hinton]{lecun2015deep}
LeCun, Y., Bengio, Y., and Hinton, G.
\newblock Deep learning.
\newblock \emph{Nature}, 2015.

\bibitem[Ledig et~al.(2016)Ledig, Theis, Husz{\'a}r, Caballero, Cunningham,
  Acosta, Aitken, Tejani, Totz, Wang, et~al.]{ledig2016photo}
Ledig, C., Theis, L., Husz{\'a}r, F., Caballero, J., Cunningham, A., Acosta,
  A., Aitken, A., Tejani, A., Totz, J., Wang, Z., et~al.
\newblock Photo-realistic single image super-resolution using a generative
  adversarial network.
\newblock \emph{arXiv preprint arXiv:1609.04802}, 2016.

\bibitem[Ling \& Okada(2006)Ling and Okada]{lapl1cost}
Ling, H. and Okada, K.
\newblock Diffusion distance for histogram comparison.
\newblock In \emph{CVPR}, 2006.

\bibitem[{Lipton} \& {Tripathi}(2017){Lipton} and {Tripathi}]{precise_recovery}
{Lipton}, Z.~C. and {Tripathi}, S.
\newblock {Precise Recovery of Latent Vectors from Generative Adversarial
  Networks}.
\newblock \emph{arXiv preprints arXiv:1702.04782}, 2017.

\bibitem[Mairal et~al.(2008)Mairal, Elad, and Sapiro]{mairal2008sparse}
Mairal, J., Elad, M., and Sapiro, G.
\newblock Sparse representation for color image restoration.
\newblock \emph{IEEE Transactions on Image Processing}, 2008.

\bibitem[Mikolov et~al.(2013)Mikolov, Chen, Corrado, and
  Dean]{mikolov2013efficient}
Mikolov, T., Chen, K., Corrado, G., and Dean, J.
\newblock Efficient estimation of word representations in vector space.
\newblock \emph{arXiv preprint arXiv:1301.3781}, 2013.

\bibitem[Nguyen et~al.(2017)Nguyen, Yosinski, Bengio, Dosovitskiy, and
  Clune]{nguyen2016ppgn}
Nguyen, A., Yosinski, J., Bengio, Y., Dosovitskiy, A., and Clune, J.
\newblock Plug \& play generative networks: Conditional iterative generation of
  images in latent space.
\newblock In \emph{CVPR}, 2017.

\bibitem[Pearson(1901)]{pca_original}
Pearson, K.
\newblock On lines and planes of closest fit to systems of points in space.
\newblock \emph{The London, Edinburgh, and Dublin Philosophical Magazine and
  Journal of Science}, 1901.

\bibitem[Portilla \& Simoncelli(2000)Portilla and Simoncelli]{Portilla2000}
Portilla, J. and Simoncelli, E.~P.
\newblock A parametric texture model based on joint statistics of complex
  wavelet coefficients.
\newblock \emph{International journal of computer vision}, 40\penalty0
  (1):\penalty0 49--70, 2000.

\bibitem[Rabiner \& Schafer(2007)Rabiner and Schafer]{RabinerS07}
Rabiner, L.~R. and Schafer, R.~W.
\newblock Introduction to digital speech processing.
\newblock \emph{Foundations and Trends in Signal Processing}, 1\penalty0
  (1/2):\penalty0 1--194, 2007.

\bibitem[{Radford} et~al.(2015){Radford}, {Metz}, and {Chintala}]{dcgan}
{Radford}, A., {Metz}, L., and {Chintala}, S.
\newblock {Unsupervised Representation Learning with Deep Convolutional
  Generative Adversarial Networks}.
\newblock \emph{arXiv preprint arXiv:1511.06434}, 2015.

\bibitem[Salimans et~al.(2016)Salimans, Goodfellow, Zaremba, Cheung, Radford,
  and Chen]{salimans2016improved}
Salimans, T., Goodfellow, I., Zaremba, W., Cheung, V., Radford, A., and Chen,
  X.
\newblock Improved techniques for training gans.
\newblock In \emph{NIPS}, 2016.

\bibitem[Santurkar et~al.(2017)Santurkar, Budden, and
  Shavit]{santurkar2017generative}
Santurkar, S., Budden, D., and Shavit, N.
\newblock Generative compression.
\newblock \emph{arXiv preprint arXiv:1703.01467}, 2017.

\bibitem[Sohl-Dickstein et~al.(2015)Sohl-Dickstein, Weiss, Maheswaranathan, and
  Ganguli]{nonequilibrium}
Sohl-Dickstein, J., Weiss, E.~A., Maheswaranathan, N., and Ganguli, S.
\newblock Deep unsupervised learning using nonequilibrium thermodynamics.
\newblock \emph{arXiv preprint arXiv:1503.03585}, 2015.

\bibitem[Stuart \& Ord(2010)Stuart and Ord]{kendall}
Stuart, A. and Ord, K.
\newblock \emph{Kendall's Advanced Theory of Statistics}.
\newblock Wiley, 2010.

\bibitem[Turk \& Pentland(1991)Turk and Pentland]{eigenfaces}
Turk, M.~A. and Pentland, A.~P.
\newblock Face recognition using eigenfaces.
\newblock In \emph{CVPR}, 1991.

\bibitem[{Ulyanov} et~al.(2017){Ulyanov}, {Vedaldi}, and
  {Lempitsky}]{adversarial_generator}
{Ulyanov}, D., {Vedaldi}, A., and {Lempitsky}, V.
\newblock {Adversarial Generator-Encoder Networks}.
\newblock \emph{arXiv preprints arXiv:1704.02304}, 2017.

\bibitem[Vincent et~al.(2008)Vincent, Larochelle, Bengio, and
  Manzagol]{vincent2008extracting}
Vincent, P., Larochelle, H., Bengio, Y., and Manzagol, P.-A.
\newblock Extracting and composing robust features with denoising autoencoders.
\newblock In \emph{ICML}, 2008.

\bibitem[Zeiler \& Fergus(2014)Zeiler and Fergus]{visualizing_cnn}
Zeiler, M.~D. and Fergus, R.
\newblock Visualizing and understanding convolutional networks.
\newblock In \emph{ECCV}, 2014.

\bibitem[{Zhang} et~al.(2016){Zhang}, {Bengio}, {Hardt}, {Recht}, and
  {Vinyals}]{understanding}
{Zhang}, C., {Bengio}, S., {Hardt}, M., {Recht}, B., and {Vinyals}, O.
\newblock {Understanding deep learning requires rethinking generalization}.
\newblock \emph{arXiv preprint arXiv:1611.03530}, 2016.

\bibitem[Zhao et~al.(2016)Zhao, Mathieu, and LeCun]{zhao2016energy}
Zhao, J., Mathieu, M., and LeCun, Y.
\newblock Energy-based generative adversarial network.
\newblock \emph{arXiv preprint arXiv:1609.03126}, 2016.

\bibitem[Zhu et~al.(2016)Zhu, Kr{\"a}henb{\"u}hl, Shechtman, and
  Efros]{zhu2016generative}
Zhu, J.-Y., Kr{\"a}henb{\"u}hl, P., Shechtman, E., and Efros, A.~A.
\newblock Generative visual manipulation on the natural image manifold.
\newblock In \emph{ECCV}, 2016.

\end{thebibliography}

\end{document}